\documentclass[letterpaper]{article}

\usepackage{natbib,alifeconf}  
\usepackage{url,hyperref,cleveref}
\usepackage{booktabs}

\usepackage{lipsum}

\usepackage{subcaption}

\newcommand\blfootnote[1]{%
  \begingroup
  \renewcommand\thefootnote{}\footnote{#1}%
  \addtocounter{footnote}{-1}%
  \endgroup
}

\title{Environmental requirements for the use of social information by artificial life agents using evolved plastic artificial neural networks.}

\author{
    Hugh Charterton$^{1*}$,
    James M. Borg$^1$,
    Anik\'{o} Ek\'{a}rt$^1$\\
    \mbox{}\\
    $^1$Aston Centre for AI Research and Application, Aston University, Birmingham, United Kingdom \\
    $^*$240403850@aston.ac.uk
} 

\begin{document}

\maketitle

\begin{abstract}
    Evolved Plastic Artificial Neural Networks (EPANNs) consist of two principal processes, the first, evolution, and the second, development and in-life learning. In the context of the origins of social learning, very few studies have been carried out using ALIFE models based on EPANN requirements. Studies in this field have usually involved an imitative teacher/pupil relationship. This, however, ignores the possibility that the observed behaviour is a consequence of social information cues rather than direct imitation or teaching.
    
    Starting with the first of the EPANN processes (evolution), a series of experiments was undertaken using artificial neural network (ANN) based agents in a variety of foraging environments to examine under what minimal environmental conditions the use of social information might have evolved, as measured by the number of generations taken to meet a specified fitness criterion. NEAT (Neuroevolution of Augmenting Topologies) was the ANN used as its evolutionary algorithm would evolve a network’s topology as well its weights. 

    Unintentionally, in the experiment there was a simple network topology based on the location of the nearest food item which enabled agents to swiftly meet the fitness criterion. With this topology, additional information, social or otherwise, was not required and could have proved to be a hindrance. However, this does indicate that for the use of social information to have evolved, it would require a greater degree of complexity in the environment to do so.
\end{abstract}

Submission type: \textbf{Full Paper}\\

Code available at: \url{https://osf.io/kugnb/files/osfstorage}
\blfootnote{\textcopyright  2026 [HUGH CHARTERTON]. Published under a Creative Commons Attribution 4.0 International (CC BY 4.0) license.}

\section{Introduction}
\citeauthor{Soltoggio2018} introduced the concept of Evolved Plastic Artificial Neural Networks (EPANNs) in 2018, which they defined as computationally based systems inspired by the evolutionary processes observed in the natural world which “employ simulated evolution in-silico to breed plastic neural networks with the aim to autonomously design and create learning systems”.

The aims, properties and evolutionary algorithmic requirements for an EPANN were outlined and EPANN-related progress to date in a variety of related fields such as plasticity \citep{soltoggio2008neural}, evolutionary robotics \citep{nitschke2012evolving}, evolved learning \citep{bullinaria2009lifetime} and neuromodulation  \citep{soltoggio2008evolutionary} reviewed. The question as to what environmental conditions, natural or computational, might promote the evolution of neural plasticity, learning and intelligence was also raised.

The final section of \citet{Soltoggio2018} reviewed potential directions for EPANNs, one such being “Incremental and social learning” where it was suggested that an area where incremental learning might find a role is in social learning where an EPANN learns about its environment through communication, observation or imitation of other EPANN instances.

Both of the examples cited by \citet{Soltoggio2018}, \citep{mcquesten1997culling,bullinaria2017imitative} used forms of imitation as the means of knowledge transfer between individuals, as have other subsequent Artificial Life investigations and social learning models \citep{jolley2016analysis,bartoli2020mechanisms,borg2011discovering,bourahla2022knowledge}, where knowledgeable individuals make that knowledge available to others based on a teacher/pupil, parent/child relationship, though that knowledge would have to, at some point, been gained through experiential learning \citep{heyes1994social,laland2004social}.

However, \citet{noble2002imitation} have questioned the perceived prevalence of imitative behaviour in nature. It was argued that what was often interpreted as imitative learning could quite easily have been the result of an observational heuristic based on social information cue. An experiment in which rhesus monkeys, which had previously shown no reaction to the presence of a snake, only exhibited a fearful response once they had heard the fearful response of a monkey who had had previous experience of snakes \citep{mineka2013social}. While this might appear imitative, it was their ability to recognize the fear cue that allowed them to subsequently respond fearfully to the presence of a snake, their heuristic in this case being “Pay attention to what others are doing or experiencing, and if the results for them appear to be good or bad then learn from this”, or more simply, watch what others are doing and do likewise.

Similarly in the case of Norway rats which, despite it being poisonous, would eat a new food they had smelt on the breath of another \citep{galef1996social}. The heuristic for the rat being “Pay attention to what others are eating and do likewise”, this being based on the observation that the other rat had not died.

The question we are looking to answer in this paper is how might this use of social information have evolved, the hypothesis being that using social information gives an evolutionary advantage resulting in improved performance in a foraging task? 

To this end we used an agent-based artificial life model which used a minimally configured EPANN as the basis for each agent rather than a pre-defined heuristic or learning algorithm. Agents of varying capabilities were given a number of foraging environments of differing environmental conditions and allowed to evolve sustainable solutions over a fixed number of generations.

Although environments and technology are not the same, these experiments explore a territory similar to that of \citet{borg2020effect} who showed that social information without learning within lifetime could be useful until insurmountably challenging environments were encountered.

\section{Technical Background}
NEAT (Neuroevolution of Augmenting Topologies) \citep{Stanley2002} was cited by Andrea Soltoggio et al. \citeyear{Soltoggio2018} as a prime example of an EPANN’s evolutionary algorithm since it managed not only the mutation and recombination of network structures (nodes, connections, layers and weighting), but also allowed diverse network structures to evolve and grow. 

However NEAT, as with other EAs, does not incorporate the plasticity required to support the evolution of an emergent learning capability and while the investigation covered in this paper uses NEAT as its evolutionary algorithm, it did so without enhancing it to include such within lifetime plasticity.

The following experiment was carried out using NEAT-python \footnote{CodeReclaimers version 1.1.0 \citep{NeatPython}} and it should be noted that there are documented differences in implementation to the original \citep{NeatPython}. It should also be noted that the reproductive process takes place at the end of a generational cycle, as defined by a number of timesteps, with the next generation of each species being derived from both a set number of its fittest members and a random selection of other individuals.

The configuration used in this work generally follows the default NEAT parameter settings with the exceptions found in table \ref{tab:Config}.
\begin{table}[!ht]
    \centering
    \begin{tabular}{p{0.3\linewidth}  p{0.6\linewidth}  p{0.2\linewidth}}
        \toprule
        Configuration Item & Description and setting\\
        \midrule
        pop\_size & Number of agents (dynamically set as per the run requirements).\\\\
        num\_inputs & Number of network inputs (see table \ref{tab:Combos} for values).\\\\
        feed\_forward & Network configuration. Set to false to enable recurrent connectivity.\\\\
        activation\_default & TANH select as the node activation function as it has a sharp response in the range $[-1,1]$ corresponding to movements along an axis, providing forward/backward, left/right movement.\\\\
        activation\_options & List of other activation functions that could be used. Set to TANH for consistency.\\\\
        fitness\_threshold & Set to 150 as derived through initial trials and based on agents’ energy value at which they were observed to be self-sustaining.\\\\
        \bottomrule
    \end{tabular}
    \caption{
        NEAT configuration values (where different from default).
    }
    \label{tab:Config}
\end{table}

\section{Experimental Setup}
A varying number of NEAT-based agents were randomly placed in a rectangular foraging  environment measuring $800 \times 600$ pixels as were a varying number of food and poison items of different ratios (see table \ref{tab:Variables} for agent and item numbers, and item ratios used).  
\begin{table}[!ht]
    \centering
    \begin{tabular}{p{0.2\linewidth}  p{0.1\linewidth}  p{0.1\linewidth}  p{0.1\linewidth}
    p{0.1\linewidth}  p{0.1\linewidth}}
        \toprule
        Environment Variables & Value a & Value b & Value c & Value d & Value e\\
        \midrule
        Agent Population  & 30 & 40 & 50 & 60 & 70\\\\
        Total Food/Poison Items  & 40 & 45 & 50 & 55 & 60\\\\
        Food/Poison Ratio  & 1:1 & 1.3:1 & 1.8:1 & 2:1 \\\\
        \bottomrule
    \end{tabular}
    \caption{
        Variable values for agent population, total number of food/poison items and the varying food/poison ratios. During a run, additional food/poison items would re-spawn if a random number between zero and one was less than 0.2 times the food/poison ratio.
    }
    \label{tab:Variables}
\end{table}

The agents were sized at $20 \times 20$ pixels and the items at $15 \times 15$ pixels. Energy and fitness had different counters so that an agent's mortality (an agent was deemed to have died if its energy value reached zero, and was removed) and its progression towards being self-sustaining could be tracked separately. If an agent overlapped an item, the item would automatically be consumed with an energy increase of $(+50)$ for food or an energy decrease of  $(-50)$ for poison. Each agent had an energy starting point of $+100$. This consumption in turn mapped to the agent’s fitness value which, starting from 0, increased or decreased by 10.

An energy loss of $0.1$ was incurred for each of the 1000 timesteps in a generation. Consequently, an agent would die at the end of a generation if it did nothing. Each movement also incurred an energy loss of $0.1$. These losses were not reflected in the fitness value since they did not reflect the agent's progress towards its fitness goal.

In addition, inspired by the bioluminescence found throughout nature, each agent would have a beacon. Configured to reflect an agent’s energy, the beacon would act as a potential source of social information from which other agents could potentially infer a strategy for foraging in the environment based on that agent’s energy health. 

An agent could therefore interact with inputs from three types of artifact, food/poison items, other agents and itself (e.g. energy). However, only information which was within an agent's area of visibility would be available to it, with visibility defined as being within a quarter of the length of the environment's diagonal. Information on each of the three artifact types would make up possible inputs to an agent's NEAT instance as described in table \ref{tab:Inputs}.  
\begin{table}[!ht]
    \centering
    \begin{tabular}{p{0.14\linewidth} p{0.55\linewidth}  p{0.14\linewidth}}
        \toprule
        Input Code & Description & \# NEAT Inputs\\
        \midrule
        FP\_XY & XY co-ordinates of the nearest visible Food/Poison item relative to agent.& 2\\\\
        FP\_IND &Indicator of whether an item is either food or poison.& 1\\\\
        AG\_EN &Normalized value (n) of agent’s own energy (e) as calculated in equation \ref{enn}
        \begin{equation}
            n=max(0.0, min(1.0,e/f))
        \label{enn}
        \end{equation}
        where $f$ refers to fitness threshold in units of energy
        & 1\\\\
        NA\_XY &XY co-ordinates of other agent relative to agent with brightest beacon as calculated using equation \ref{oab} & 2\\\\
        NA\_B &Beacon value ($b'$ in equation \ref{oab}).\newline where $'$ refers to other agent\newline  $e'$ refers to other agent's energy\newline $x'$ refers to other agent's x axis co-ordinate\newline $y'$ refers to other agent's y axis co-ordinate:
        \begin{equation}
            b' = \frac{e'}{sqrt{((y'-y)^2+(x'-x)^2)}}
        \label{oab}
        \end{equation}
        & 1\\
        \bottomrule
    \end{tabular}
    \caption{
        Inputs, descriptions and the number of NEAT inputs required.
    }
    \label{tab:Inputs}
\end{table}

As the information provided by another agent did not give information about a potential energy source, it was deemed to provide indirect information, while information pertaining to a potential energy source was deemed to provide direct information, giving each agent three types of possible interaction with the environment as shown in table \ref{tab:Iacts}.
\begin{table}[!ht]
    \centering
    \begin{tabular}{p{0.15\linewidth}  p{0.3\linewidth}  p{0.45\linewidth}}
        \toprule
        Interaction Id&direct/indirect&Description\\
        \midrule
        1&direct&Information about item and the agent its self.\\\\
        2&indirect&Information about another agent.\\\\
        3&direct \& indirect&Combined direct and indirect information\\\\
        \bottomrule
    \end{tabular}
    \caption{
        Direct/Indirect Interaction types.
    }
    \label{tab:Iacts}
\end{table}
Each interaction would be composed of appropriate combinations of NEAT inputs (See table \ref{tab:Combos}) with each combination forming the basis for a set of runs across each combination of the environment configuration variables.
\begin{table}[!ht]
    \centering
    \begin{tabular}{p{0.15\linewidth}  p{0.1\linewidth}  p{0.45\linewidth}  p{0.15\linewidth}}
        \toprule
        NEAT input combination ID & Inter-action ID & Input codes & Total \newline Inputs\\
        \midrule
        1 & 1 & FP\_XY & 2\\
        2 & 1 & FP\_XY, FP\_IND & 3\\
        3 & 1 & FP\_XY, AG\_EN & 3\\
        4 & 1 & FP\_XY, FP\_IND, AG\_EN & 4\\
        5 & 2 & NA\_XY, NA\_B & 3\\
        6 & 2 & NA\_XY, NA\_B, AG\_EN & 4\\
        7 & 3 & FP\_XY, NA\_XY, NA\_B & 5\\
        8 & 3 & FP\_XY, FP\_IND, NA\_XY, NA\_B & 6\\
        9 & 3 & FP\_XY, AG\_EN, NA\_XY, NA\_B & 6\\
        10 & 3 & FP\_XY, FP\_IND, AG\_EN, NA\_XY, NA\_B & 7\\
        \bottomrule
    \end{tabular}
    \caption{
        NEAT input combinations, grouped by their type of interaction with the environment and their total number of inputs to a NEAT neural network. Please note that combination 6 has been classified as indirect since the agent's energy is viewed as complementary to the information about another agent.    
    }
    \label{tab:Combos}
\end{table}

All agent NEAT networks had two outputs which were used to update its position in the environment relative to its current position at the end of each timestep, as shown in table \ref{tab:Outputs}.
\begin{table}[!ht]
    \centering
    \begin{tabular}{p{0.2\linewidth}  p{0.5\linewidth}  p{0.2\linewidth}}
        \toprule
        Output code & Description & \# Outputs\\
        \midrule
        X & Movement along X axis& 1\\
        Y & Movement along Y axis& 1\\
        \bottomrule
    \end{tabular}
    \caption{
        NEAT neural network output values. As NEAT would return values between -1 and 1, a speed value was used to map the returned value to give the number of pixels to move. The speed value was set to 5.
    }
    \label{tab:Outputs}
\end{table}

20 runs over 120 generations, each of 1000 timesteps were executed for each input code combination across each combination of the environment configuration variables, a total of $20,000$ runs.

At the end of each generation, the initial variable values, outcome with respect to fitness threshold and the network structure of the fittest agent were logged for subsequent analysis.

\section{Results}
\begin{figure}[!ht]
    \centering
    \includegraphics[width=3.1in,angle=0]{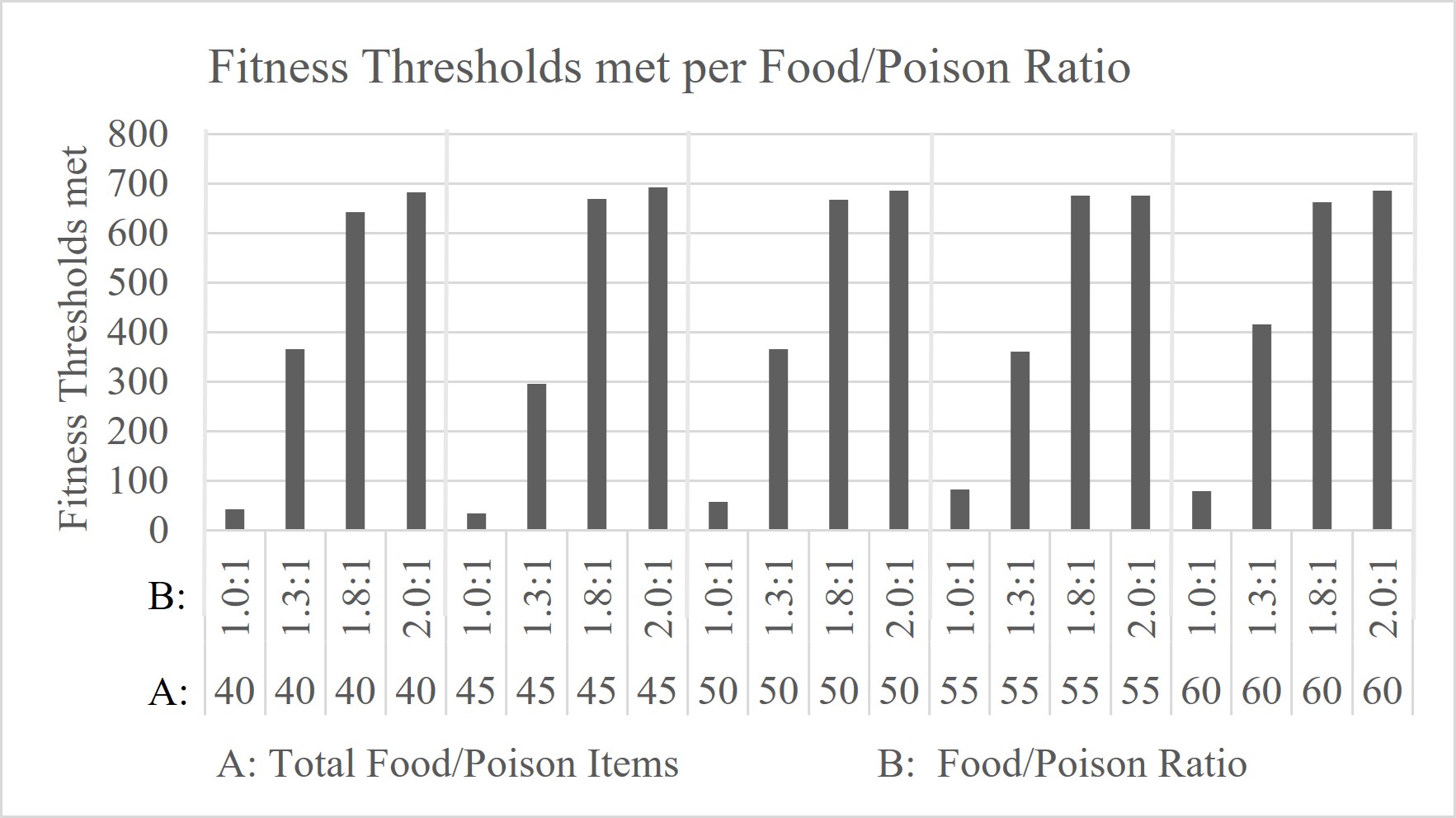}
    \caption{
        Number of fitness thresholds met for each ratio of food/poison items for each initial total of food/poison items in the environment across all populations and all input code combinations.
    }
    \label{fig1}
\end{figure}

The effectiveness of any environmental configuration was measured in terms of the number of times the fitness threshold was met per configuration. As might be expected, the number of thresholds being met increased as the ratio of food to poison items increased with similar numbers across all food/poison item totals as shown in figure \ref{fig1}.

\begin{figure}[!ht]
    \centering
    \includegraphics[width=3.1in,angle=0]{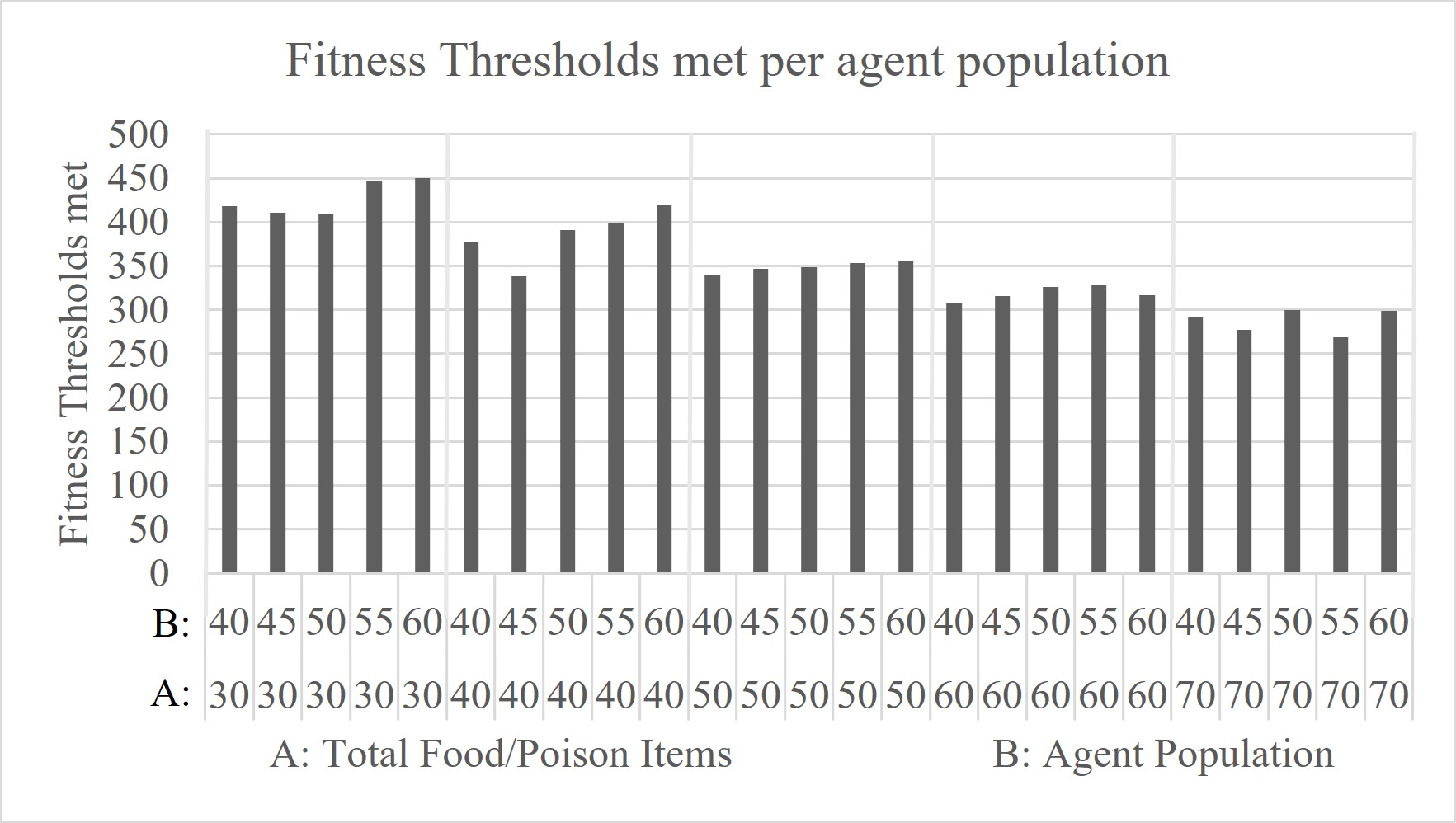}
    \caption{
        Number of fitness thresholds met across the range of population values for each initial total of food/poison items in the environment across all food/poison ratios and all input code combinations.
    }
    \label{fig2}
\end{figure}

Variations in the population had no discernible impact other than a steady overall decrease in effectiveness as the population grew in spite of the higher volumes of food/poison items as shown in figure \ref{fig2}.

\begin{figure}[!ht]
    \centering
    \includegraphics[width=3.1in,angle=0]{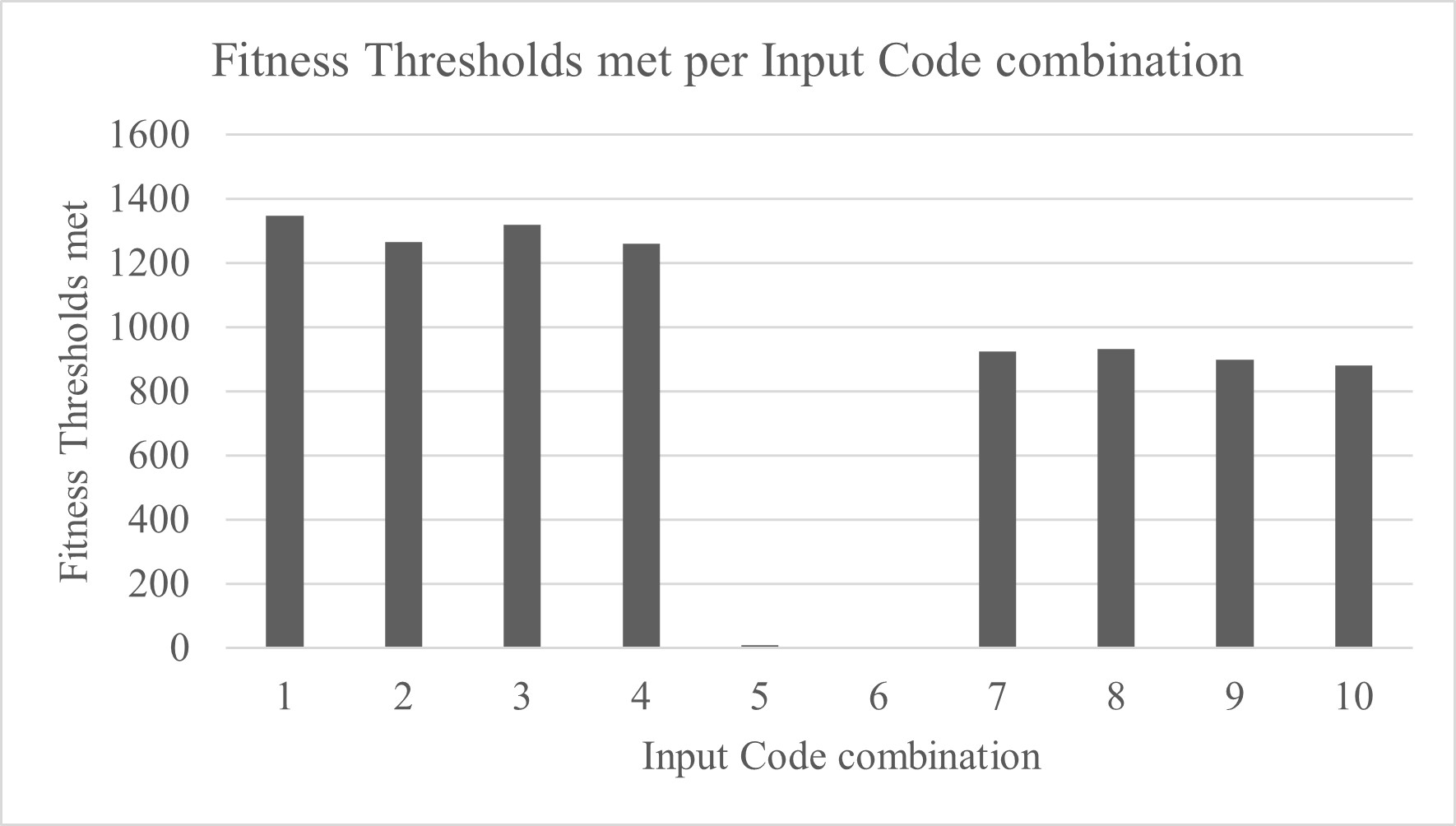}
    \caption{
        Fitness thresholds meeting the termination criteria for each NEAT input combination. For NEAT input combinations, see table \ref{tab:Combos}.
    }
    \label{fig3}
\end{figure}

When reviewing the fitness threshold results by NEAT input combination alone, they divided into their interaction types with those combinations which interacted directly with the environment (NEAT input combinations 1 – 4) achieving the greatest number of thresholds met while those interacting directly and indirectly (NEAT input combinations 7 – 10) came second. Those which only interacted indirectly with the environment (NEAT input combinations 5 \& 6) hardly ever met the required thresholds, as can be seen in figure \ref{fig3}.

A subsequent set of 50 runs each of 1000 generations was carried out for the indirect NEAT input combinations over a population of 40 agents with a food/poison item total of 55 and a food/poison ratio of $1.3:1$, with the aim of looking into the possibility that additional generations might allow a higher number of thresholds to be met. However, no thresholds were met at all. 

\begin{figure}[!ht]
    \centering
    \includegraphics[width=3.1in,angle=0]{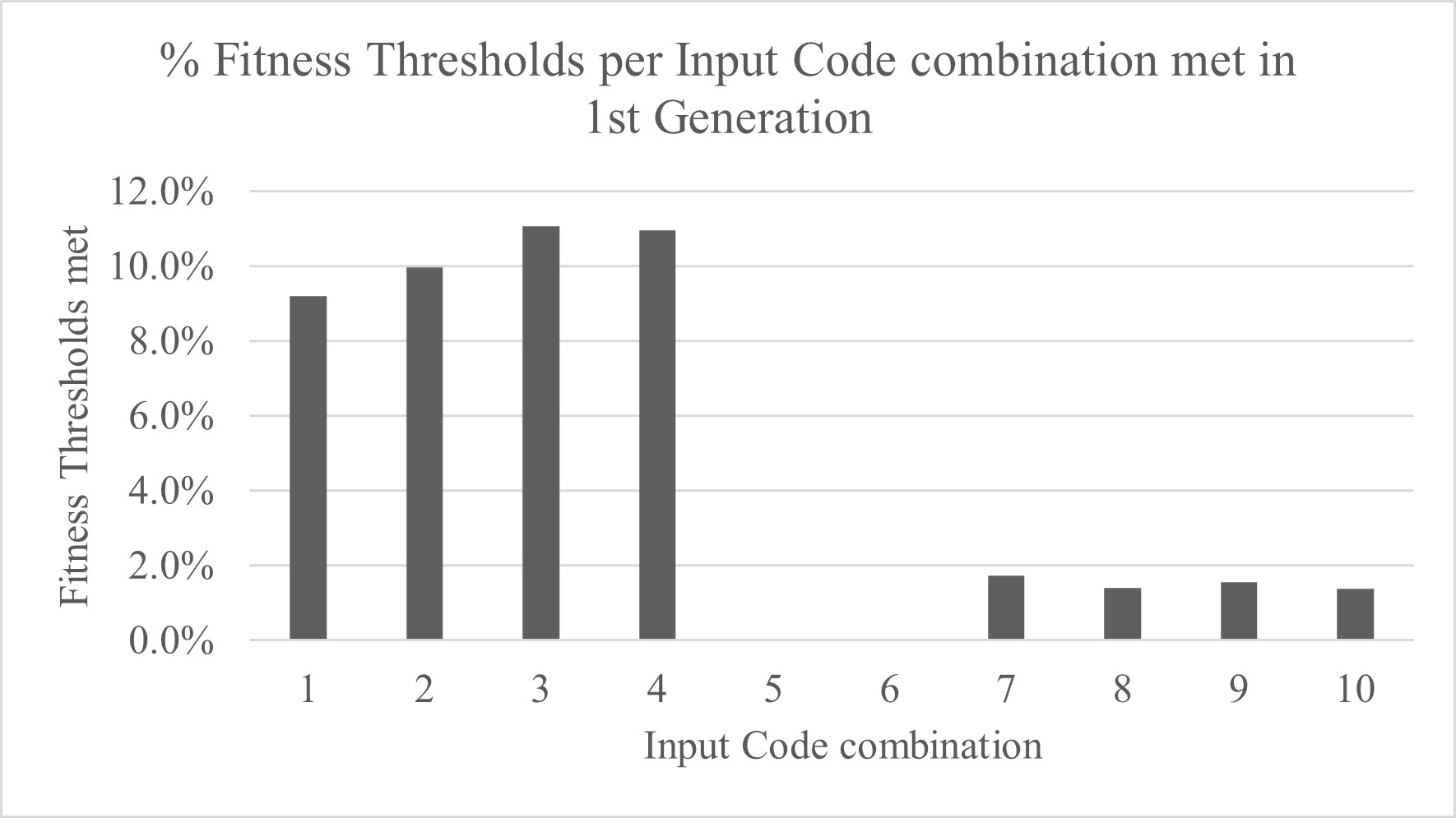}
    \caption{
        Percentage of thresholds met in their first generation out of the total number of thresholds met per input code combination. where 9-11\% of each of the direct NEAT input combinations meets the threshold in their first generation, a strong indicator that this environment might have been too simple.
        Those with combined direct and indirect NEAT input combinations meet the threshold approximately 1\% of the time in their first generation, while the indirect combinations never reached the threshold in their first generation.
        For details on each input code combination, see table \ref{tab:Combos}. 
    }
    \label{fig4}
\end{figure}

\begin{figure}[!ht]
    \centering
    \begin{subfigure}[b]{0.2\textwidth}
        \includegraphics[width=1.5in,angle=0]{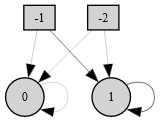}
        \caption{NEAT network topology for NEAT input combination 1 (as defined in table \ref{tab:Combos})}
        \label{image5a}
    \end{subfigure}
    \hfill
    \begin{subfigure}[b]{0.2\textwidth}
        \includegraphics[width=1.5in,angle=0]{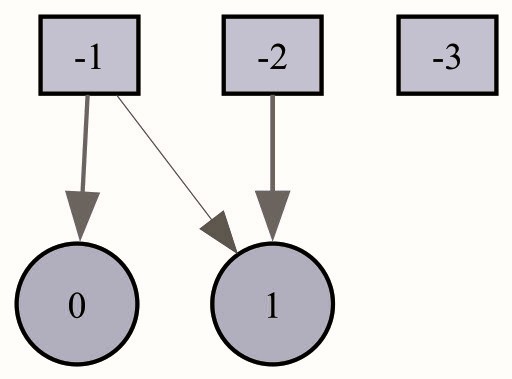}
        \caption{NEAT network topology for NEAT input combination 3 (as defined in table \ref{tab:Combos})}
        \label{image5b}
    \end{subfigure}
    \caption{Examples of first generation NEAT topologies which reached the fitness threshold in which NEAT inputs connect directly to the outputs with no hidden layers. Both examples had initial agent populations of 30, initial number of food/poison items of 55 with a ratio of $2:1$.}
    \label{fig5}
\end{figure}

One possible reason that the direct NEAT input combinations performed best might have been that there was a simple network configuration based only on the co-ordinates of the nearest food/poison item which provided the agent with a survival solution such that it could ignore other inputs and information be it additional direct environmental information or indirect social information. For instance, between 9 and 11 percentage of each of the direct input code combinations meeting the required threshold did so in the first generation as shown in figure \ref{fig4}. Reviewing the network configurations from the first generation threshold-meeting outcomes showed that only a network topology featuring direct connections between the FP\_XY input, as defined in table \ref{tab:Inputs}, and the two output nodes was required, as illustrated in the examples in figure \ref{fig5}.

\begin{figure}[!ht]
    \centering
    \includegraphics[width=0.5\textwidth]{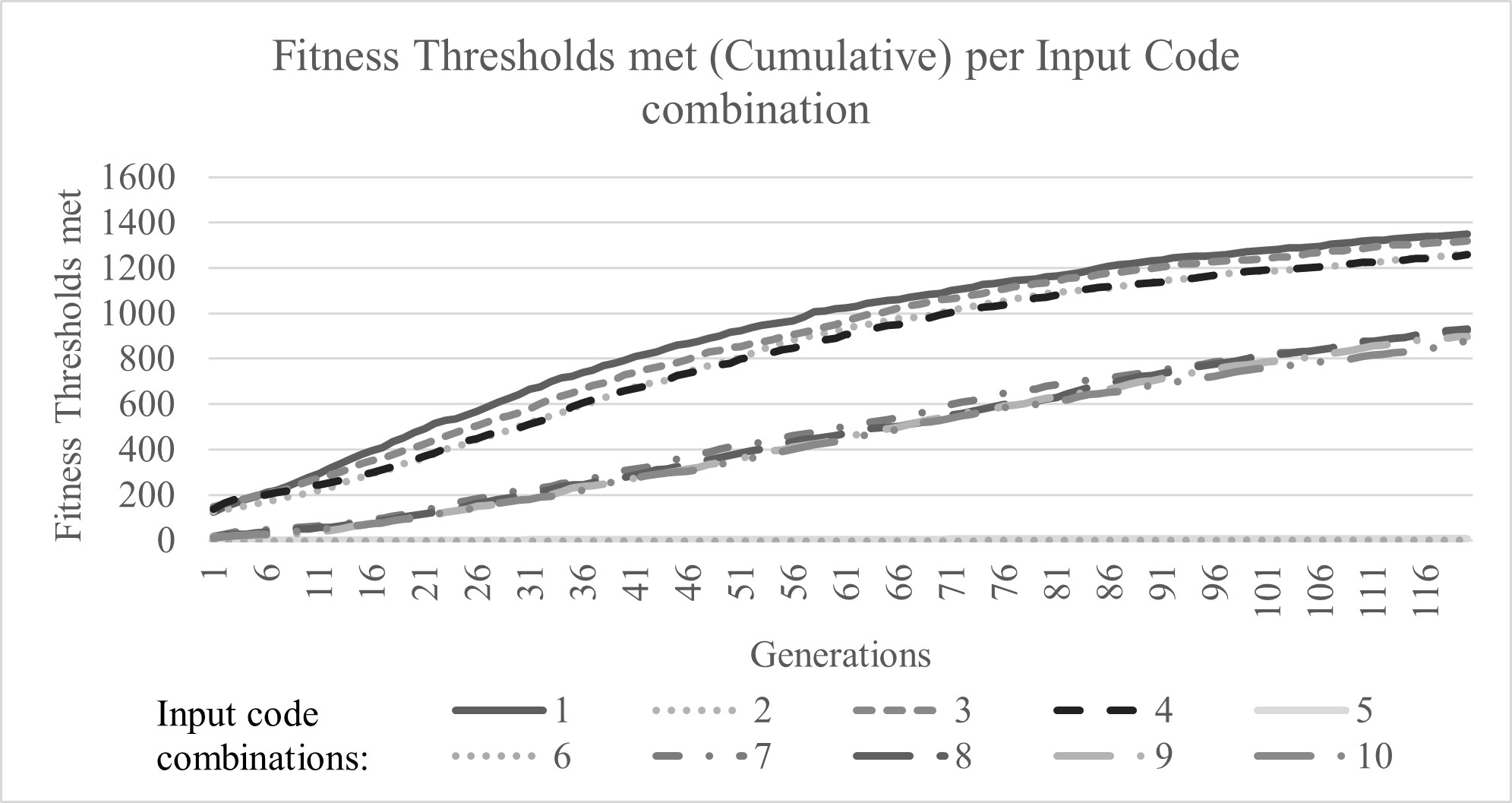}
    \caption{
        Cumulative number of fitness thresholds met per input code combination as defined in table \ref{tab:Combos}.
    }
    \label{fig6}
\end{figure}

Comparing the total number of runs where the fitness threshold was met for each NEAT input combination across all generations and for each population number, total food/poison items and their ratios, shows a distinct split between the interaction types as illustrated in figure \ref{fig6} which shows the cumulative total of thresholds met for each of the input code combinations over 120 generations for each population, number of food/poison items and food/poison ratios.

While the NEAT input combinations for each interaction type follow the same trajectory, it was noted that the leading combination for each interaction type was the one with the least number of NEAT network inputs.

Figure \ref{fig6} also shows that the number of thresholds being met by the direct NEAT input combinations initially increases at a faster rate than those for the combined direct and indirect combinations, while the indirect combinations hardly register at all.

\section{Conclusion}
This study has been carried out with only half an EPANN, in that only an evolutionary component, in the form of NEAT, was deployed, in runs of a short generational time span, across environments with a varying number of agents and food/poison items of differing ratios. The only environmental change during a generational lifetime was the reduction and subsequent re-spawning of items due to agent consumption.

Given that 10\% of the direct input combinations had a solution before NEAT’s evolutionary algorithm was invoked suggests that the environment was too simple to provide evidence as to whether social information improves performance in an EPANN/NEAT based foraging environment. Where this was not the case, the number of thresholds being met decreased as the number of NEAT inputs increases, as can be seen in figure \ref{fig3}. It is therefore unlikely that the addition of an in-life learning capability would, in this instance, have made any difference. However, it should be recognized that the availability of simple solutions should not automatically be seen adversely as they can provide efficient solutions as demonstrated by \citet{seth1998evolving}.   

One variable which was not altered was that of the impact of food/poison on agents. Varying this value, along with other environment changes as listed below might increase the environment’s complexity such that, on its own, the nearest item’s co-ordinates no longer provided an easy solution. 
\begin{itemize}
  \item Have an agent decide whether to consume an item or not.
  \item Have agents react when they encounter another agent, currently they ignore each other.
  \item Vary the impact of food/poison items on agents.
  \item Have food/poison items distributed in clusters.
  \item Have food/poison items which are not consumed in one timestep.
  \item Have seasonal item availability over a longer generational timespan.
\end{itemize}
Along with additional logging of movement and proximity data of other agents, it might then be possible for the beneficial use of a social information input to evolve and be recognized. A further step would be to replicate the \citet{borg2020effect} environment using NEAT and make a comparison with the original results.

Social information has been defined as “information derived from the behaviours, actions, cues or signals of other agents” \citep{borg2018emergence} and as such the beacon used in the experiment would qualify as providing social information. However, since the beacon is solely a reflection of an agent’s energy level over which the agent has no control, perhaps it better corresponds to being defined as public information since it is “inadvertent social information” \citep{danchin2004public}.

If, however, the agent manipulated its beacon value before it was visible to other agents, then it could be deemed social information, since the agent would have control over what value, if any, the beacon made available. One way this could be achieved would be through an additional NEAT output which would modulate the beacon value before it became available to other agents.

The corollary of a modulated beacon would be a capability to decode a beacon value which could be provided through neuromodulated \citep{soltoggio2007evolving,Barnes2020} within-lifetime learning, the other half of an EPANN.

While the question of how the use of social information might have evolved has not been answered in this experiment, it has shown that it is unlikely to have emerged if other simpler survival mechanisms are available. The requisite environmental complexity remains to be explored.


\footnotesize
\bibliographystyle{apalike}
\bibliography{ALIFE26} 

\end{document}